\documentclass[letterpaper, 10 pt, conference]{ieeeconf}  

\IEEEoverridecommandlockouts

\usepackage{amsmath,amssymb,amsfonts}
\usepackage{algorithmic}
\usepackage{graphicx}
\usepackage{textcomp}
\usepackage{xcolor}
\usepackage[backend=biber, style=ieee]{biblatex}
\usepackage{mdframed}
\usepackage{biblatex2bibitem}
\usepackage{flushend}
\title{\LARGE \bf
Explaining Agent Behavior with Large Language Models
}

\author{Xijia Zhang$^{1, 2}$, Yue Guo$^{3}$, Simon Stepputtis$^{3}$, Katia Sycara$^{3}$, and Joseph Campbell$^{3}$
\thanks{$^{1}$University of Michigan
        {\tt\small ponyz@umich.edu}}%
\thanks{$^{2}$Shanghai Jiao Tong University {\tt\small zhangxijia@sjtu.edu.cn}}%
\thanks{$^{3}$Carnegie Mellon University
        {\tt\small \{yueguo, sstepput, sycara, jacampbe\}@andrew.cmu.edu}}%
}

\newmdtheoremenv{def_box}{Def.}

\begin{document}

\maketitle
\thispagestyle{empty}
\pagestyle{empty}

\begin{abstract}

Intelligent agents such as robots are increasingly deployed in real-world, safety-critical settings.
It is vital that these agents are able to explain the reasoning behind their decisions to human counterparts, however, their behavior is often produced by uninterpretable models such as deep neural networks.
We propose an approach to generate natural language explanations for an agent's behavior based only on observations of states and actions, agnostic to the underlying model representation.
We show how a compact representation of the agent's behavior can be learned and used to produce plausible explanations with minimal hallucination while affording user interaction with a pre-trained large language model.
Through user studies and empirical experiments, we show that our approach generates explanations as helpful as those generated by a human domain expert while enabling beneficial interactions such as clarification and counterfactual queries.

\end{abstract}

\section{Introduction}

Rapid advances in artificial intelligence and machine learning have led to an increase in the deployment of robots and other embodied agents in real-world, safety-critical settings~\cite{sun2020scalability, fatima2017survey, boukerche2020machine, Zhou2023, li2023large}.
As such, it is vital that practitioners -- who may be laypeople that lack domain expertise or knowledge of machine learning -- are able to query such agents for explanations regarding \textit{why} a particular prediction has been made -- broadly referred to as explainable AI~\cite{amir2019summarizing, policy_summarization_review, gunning2019xai}.
While progress has been made in this area, prior works tend to focus on explaining agent behavior in terms of rules~\cite{johnson1994agents}, vision-based cues~\cite{cruz2021interactive, mishra2022not}, semantic concepts~\cite{zabounidis2023concept}, or trajectories~\cite{guo2021edge, campbell2023introspective}.
However, it has been shown that laypeople benefit from natural language explanations~\cite{mariotti2020towards, alonso2017exploratory} since they do not require specialized knowledge to understand~\cite{wang2019verbal}, leverage human affinity for verbal communication, and increase trust under uncertainty~\cite{gkatzia2016natural}.

In this work, \textbf{we seek to develop a framework to generate natural language explanations of an agent's behavior given only observations of states and actions}.
By assuming access to only behavioral observations, we are able to explain behavior produced by \textit{any} agent policy, including deep neural networks.
Unlike prior methods which exhibit limited expressivity due to utilizing language templates~\cite{hayes2017improving, kasenberg2019generating, wang2019verbal} or assume access to a large dataset of human-generated explanations~\cite{ehsan2019automated, liu2023novel}, we propose an approach in which large language models (LLMs) can be used to generate free-form natural language explanations in a few-shot manner.
While LLMs have shown considerable zero-shot task performance and are well-suited to generating natural language explanations~\cite{wiegreffe2021reframing, marasovic2021few, li2022explanations}, they are typically applied to commonsense reasoning as opposed to explaining model behavior, and are prone to hallucination -- a well-known phenomenon in which false information is presented as fact~\cite{mckenna2023sources}. It is an open question as to how LLMs can be conditioned on an agent's behavior in order to generate plausible explanations while avoiding hallucination.

Our solution, and core algorithmic contribution, is the introduction of a \textit{behavior representation} (BR), in which we distill an agent's policy into a locally interpretable model that can be directly injected into a text prompt and reasoned with, without requiring fine-tuning.
A behavior representation acts as a compact representation of an agent's behavior around a specific state and indicates what features the agent considers important when making a decision.
We show that by constraining an LLM to reason about agent behavior in terms of a behavior representation, we are able to greatly reduce hallucination compared to alternative approaches while generating informative and plausible explanations.
An additional benefit of our approach is that it enables \textit{interactive} explanations; that is, the user can issue follow-up queries such as clarification or counterfactual questions.
This is particularly valuable, as explanations are social interactions conditioned on a person's own beliefs and knowledge~\cite{miller2019explanation} and thus, are highly individual and may require additional clarification to be comprehensible and convincing~\cite{kass1988need}.

\begin{figure*}[ht]
\centerline{\includegraphics[width=0.95\textwidth]{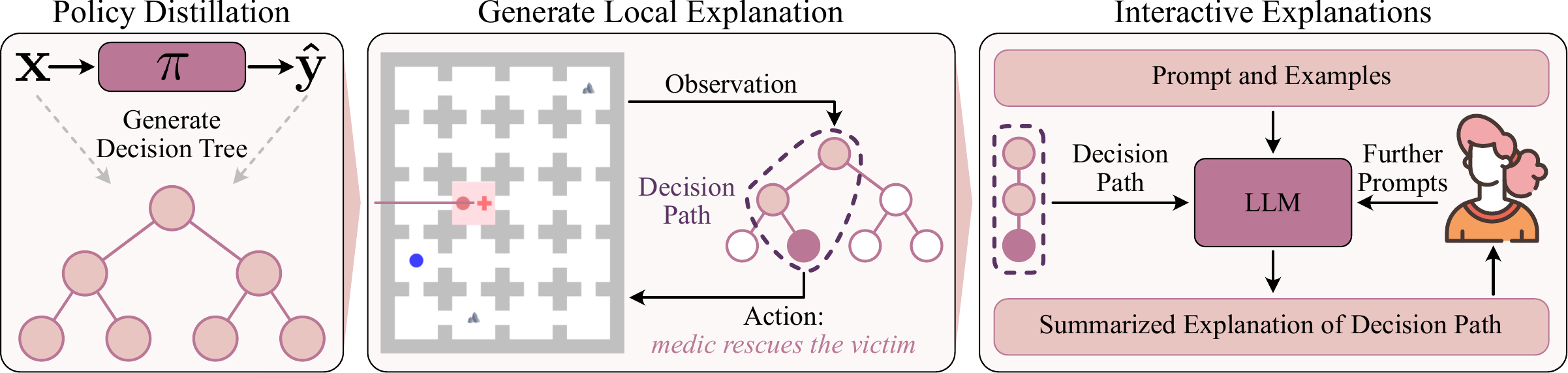}}
\caption{Overview of our three-step pipeline to explain policy actions: left: A black-box policy is distilled into a decision tree; middle: a decision path is extracted from the tree for a given state which contains a set of decision rules used to derive the associated action; right: we utilize an LLM to generate an easily understandable natural language explanation given the decision path. Lastly, a user can ask further clarification questions in an interactive manner. }
\label{usar_overview}
\end{figure*}

Our approach is a three-stage process (see Figure~\ref{usar_overview}) in which we, 1) distill an agent policy into a decision tree, 2) extract a decision path from the tree for a given state which serves as our local \textit{behavior representation}, and 3) transform the decision path into a textual representation and inject it into pre-trained LLM via in-context learning~\cite{brown2020language} to produce a natural language explanation.
In this work we show how our framework can be applied to multi-agent reinforcement learning (MARL) policies -- a particularly relevant setting given the complex dynamics and decision-making resulting from agent-agent interactions.
Through a series of participant studies, we show that, a) our approach generates model-agnostic explanations that are significantly preferred by laypeople over baseline methods, and are preferred at least as much as those generated by a human domain expert; b) when an agent policy is sub-optimal, participants find the ability to interact with our explanations helpful and beneficial; and c) our approach yields explanations with significantly fewer hallucinations than alternative methods of encoding agent behavior.

\section{Related Work}

\textbf{Explainable Agent Policies}: %
Many works attempt to explain agent behavior through the use of a simplified but interpretable model that closely mimics the original policy~\cite{puiutta2020explainable, verma2018programmatically, liu2019toward, shu2017hierarchical}, a technique which has long been studied in the field of supervised learning~\cite{ribeiro2016should, bhagat2023sample}.
Although approaches that directly utilize inherently interpretable models with limited complexity during the training phase~\cite{du2019techniques, Stepputtis2022, campbell2020learning} exist, many researchers avoid sacrificing model accuracy for interpretability.
We follow a similar approach~\cite{guo2023explainable} in this work, in which we leverage a distilled interpretable model to gain insight into how the agent's policy reasons.

\textbf{Natural Language Explanations}: %
Outside of explaining agent behavior, natural language explanations have received considerable attention in natural language processing areas such as commonsense reasoning~\cite{marasovic_natural_2020, rajani_explain_2019} and natural language inference~\cite{prasad_what_2021}.
Unlike our setting in which we desire to explain a given \textit{model's behavior}, these methods attempt to produce an explanation purely with respect to the given input and domain knowledge, e.g., whether a given premise supports a hypothesis in the case of natural language inference~\cite{camburu2018snli}.
Although self-explaining models~\cite{marasovic2021few, majumder_knowledge-grounded_2022, hu2023thought} are conceptually similar to our goal, we desire a model-agnostic approach with respect to the agent's policy and thus seek to explain the agent's behavior with a separate model.
While recent works have investigated the usage of LLMs in explaining another model's behavior by reasoning directly over the latent representation~\cite{bills2023language}, this approach has yielded limited success thus far and motivates the usage of an intermediate behavior representation in our work.

\section{Language Explanations for Agent Behavior}
\label{sec:inle}

We introduce a framework for generating natural language explanations for an agent from \textit{only} observations of states and actions.
Our approach consists of three steps: 1) we distill the agent's policy into a decision tree, 2) we generate a behavior representation from the decision tree, and 3) we query an LLM for an explanation given the behavior representation.
We note that step 1 only needs to be performed once for a particular agent, while steps 2 and 3 are performed each time an explanation is requested.
We make no assumptions about the agent's underlying policy such that our method is model agnostic; explanations can be generated for any model for which we can sample trajectories.
Formally, we examine the setting in which an agent makes an observation of an environment and executes an action at each time step.

\noindent\textbf{Notation}: An agent observes environment state $s_t$ at discrete timestep $t$, performs action $a_t$, and receives the next state $s_{t+1}$.
Actions are sampled according to the policy $\pi(a_t | s_t)$ and a trajectory consists of a sequence of state-action pairs $\tau = s_0, a_0, s_1, a_1, \dots, s_t, a_t$.

\subsection{Distilling a Decision Tree}

Our first step is to distill the agent's underlying policy into a decision tree, which acts as an interpretable \textit{surrogate}.
The decision tree is intended to faithfully replicate the agent's policy while being interpretable, such that we can extract a behavior representation from it.
Given an agent policy $\pi$, we sample $N$ rollouts $\tau_0, \dots, \tau_{N-1}$ with which we learn a decision tree policy.
While decision trees are simpler than other methods such as deep neural networks (DNNs), it has been shown that they are still capable of learning reasonably complex policies~\cite{viper}.
Intuitively, DNNs often achieve state-of-the-art performance not because their representational capacity is larger than other models, but because they are easier to regularize and thus train~\cite{ba2014deep}.
However, distillation is a technique that can be leveraged to distill the knowledge contained within a DNN into a more interpretable surrogate model~\cite{hinton2015distilling, frosst2017distilling, bastani2017interpretability}.

\subsection{Behavior Representation Generation}

Given our surrogate decision tree, the next step is to generate a behavior representation for the agent given a particular state-action pair that we wish to explain.
The behavior representation is a locally interpretable model of the agent's behavior at a particular point in the state space.
We create a behavior representation by extracting the decision path from our learned decision tree corresponding to the given state.
Intuitively, this decision path represents the decision rules that the agent may have reasoned with -- we say \textit{may} because the surrogate decision tree is only an approximation of the agent's true reasoning process -- when taking the observed action.
The decision path is an ideal behavior representation since we can algorithmically translate it to natural languge for injection into an LLM prompt without requiring any additional fine-tuning.

\subsection{In-Context Learning with Behavior Representations}

The last step in our approach is to define a prompt that constrains the LLM to reason about agent behavior with respect to a given behavior representation.
Our prompt consists of four parts: a) a concise description of the environment the agent is operating in, e.g., state and action descriptions, b) a description of what information the behavior representation conveys, c) in-context learning examples, and d) the behavior representation and action that we wish to explain.
An example of this prompt is shown in Fig.~\ref{fig:chat}.
All parts except for (d) are pre-defined ahead of time and remain constant for all queries, while our framework provides a mechanism for automatically constructing (d).
Thus, our system can be queried for explanations with no input required by the user unless they wish to interact and submit follow-up queries.

\subsection{Interactive Explanations}

Because our system leverages an LLM to generate explanations, it facilitates follow-up interactions that are conditioned on the preceding behavior representation and explanation.
As we show in Sec.~\ref{sec:results}, this is particularly valuable when the agent policy is sub-optimal, and the agent's actions are not aligned with the user's expectations.
Two common types of interactions are clarification questions, e.g., ``Why did the agent not consider feature X when making its decision?'', and counterfactual questions, e.g., ``What if feature Y were present instead of feature X?''.

\section{Participant Study and Analysis}
\label{sec:results}

\begin{figure}
    \centering
    \includegraphics[width=1\linewidth]{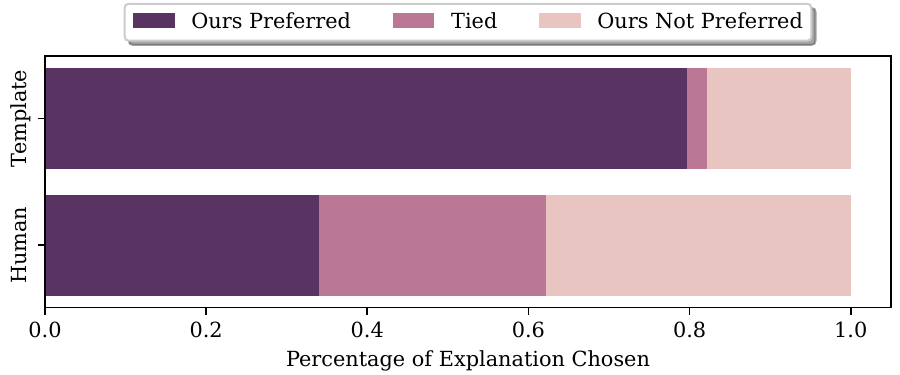}
    \caption{Participant preference when presented with two explanations and asked to choose which is most helpful to understand agent behavior: Ours vs. Template (top) and Ours vs. Human (bottom) over 320 responses.}
    \label{fig:user_preference}
\end{figure}

We conduct a series of participant studies and empirical experiments in order to answer the following questions:
\begin{itemize}
    \item Does our approach generate plausible and useful explanations?
    \item Is the user interaction enabled by our approach helpful?
    \item Is our approach able to minimize hallucination?
\end{itemize} %
We evaluate our task in a multi-agent Urban Search and Rescue (USAR) environment, implemented as a 2D grid spanning 4 $\times$ 5 rooms in which each room can contain rubble, a victim, or both (note that victims can be hidden under rubble).
The goal of our two distinct agents, an engineer, who can remove rubble, and a medic, who can heal victims, is to rescue all victims scattered throughout the rooms.
At each step, agents can take a navigation action to explore previously unexplored rooms, move through them, or conduct their unique action of removing rubble (engineer) or rescuing a victim (medic).
All experiments utilizing an LLM use OpenAI's GPT-4 model with no additional fine-tuning~\cite{gpt4}.

\subsection{Does Our Approach Produce Useful Explanations?}

We conduct an IRB-approved human-subject study with 32 participants in which we present ten environment states and actions -- 5 each for the medic and engineer -- to each participant alongside explanations generated by a) our proposed framework, b) an explanation generated via language templates directly from the behavior representation, and c) a human-generated explanation.
The explanations are presented as two pairwise decisions and each participant is asked to choose which explanation in each pair is most helpful in understanding the agent's action.

\begin{figure}
    \centering
    \includegraphics[width=1\linewidth]{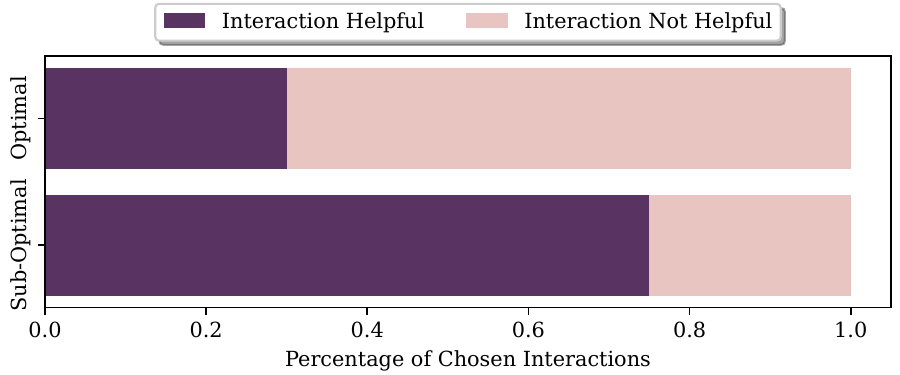}
    \caption{Helpfulness of interaction after presented with an explanation with respect to optimal and sub-optimal policies across 20 responses.}
    \label{fig:interaction_preference}
\end{figure}

\begin{figure}
    \centering
    \includegraphics[width=1\linewidth]{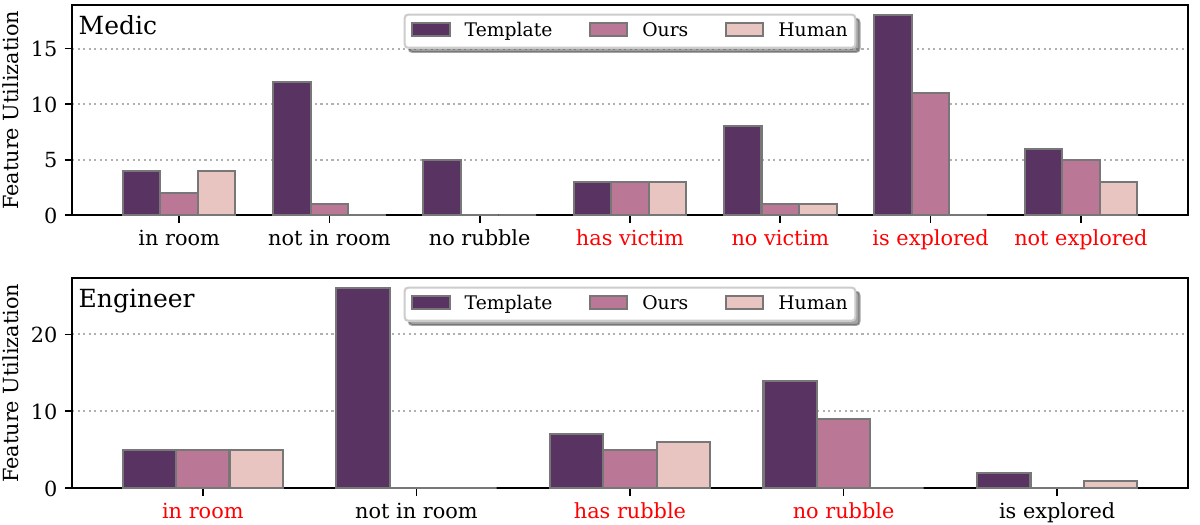}
    \caption{Frequency of selected features when explaining actions. Features highlighted in red were part of the examples provided in the LLM's prompt.}
    \label{fig:feature_selection}
\end{figure}

The results are shown in Fig.~\ref{fig:user_preference} from which we can draw two conclusions: participants significantly ($p<0.05$ with one-tailed binomial test) prefer the explanation generated by our approach over the baseline generated from language templates; and participants do \textit{not} prefer the human-generated explanations over those generated by our approach.
Analysis over the features which are referenced in each of the explanations (Fig.~\ref{fig:feature_selection}) reveals that our explanations and the human-generated explanations tend to emphasize the same features, while the template-generated explanations often refer to features that may be less helpful, e.g., ``not in room''.
From these results we reason that our explanations are indeed useful to participants.

\subsection{Is User Interaction Helpful?}

\begin{figure}
    \centering
    \includegraphics[width=1\linewidth]{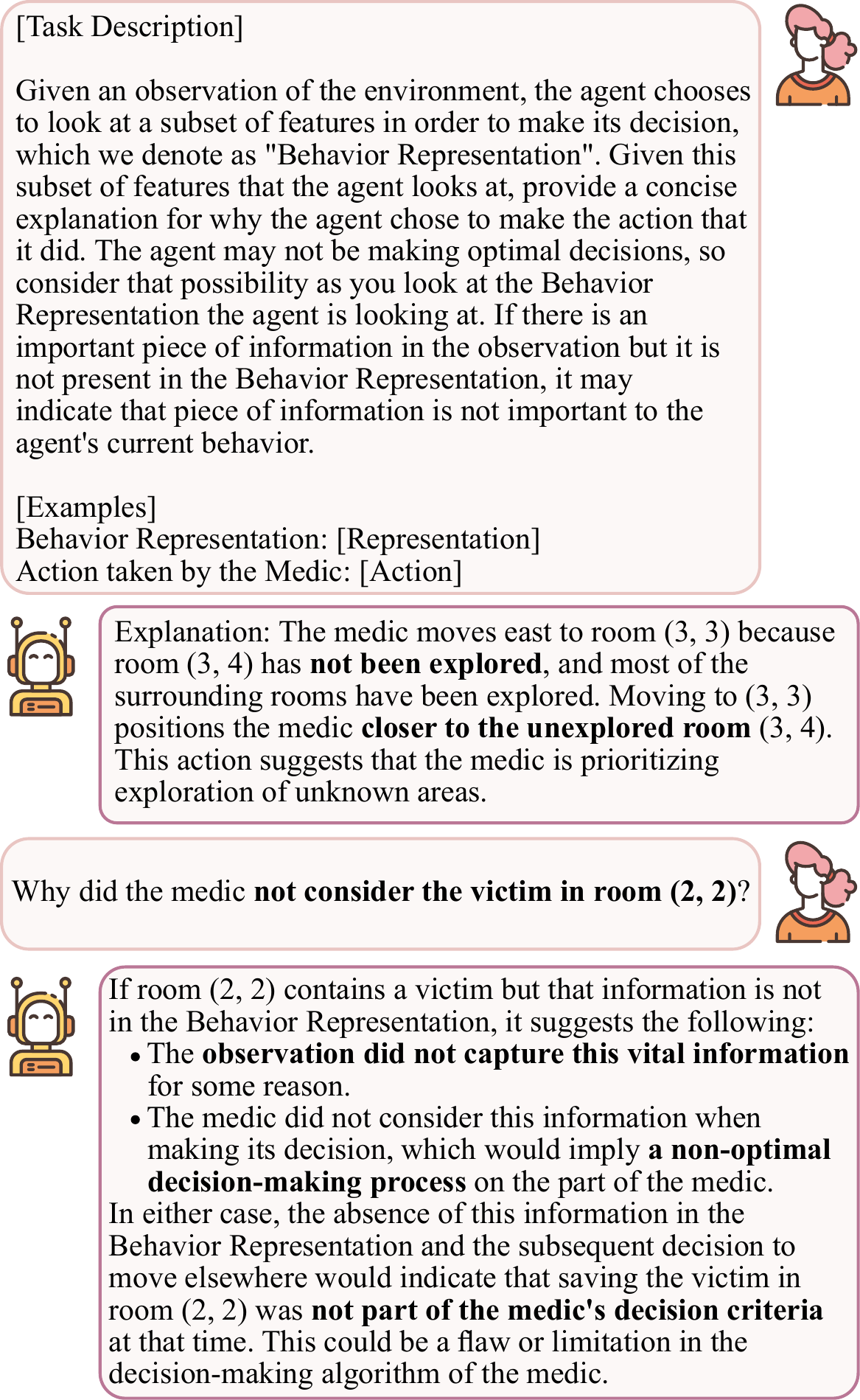}
    \caption{Interactive Conversation: The user is asking more clarification questions about the initial explanation generated by our method.}
    \label{fig:chat}
\end{figure}

We conduct a follow-up IRB-approved human-subject study with 5 participants in which we present 4 environment states, actions, and explanations generated by our framework and offer the user to further interact with the LLM via a chat window.
Participants were asked to indicate whether they found the ability to interact helpful in understanding the agent's action.
The results are shown in Fig.~\ref{fig:interaction_preference} and resulted in an interesting insight.
When the agent's policy was optimal and the action aligned with the participant's expectations, interactions were largely found as not helpful; the initial explanation was sufficient for most participants.
However, when the agent's policy was sub-optimal -- which we achieved by creating a policy in which the agents explored all rooms first before rescuing victims or removing rubble -- the participants found the ability to interact with the explanation and ask follow-up clarification questions helpful.
An example of such an interaction is shown in Fig.~\ref{fig:chat}.

\begin{figure}
    \centering
    \includegraphics[width=1\linewidth]{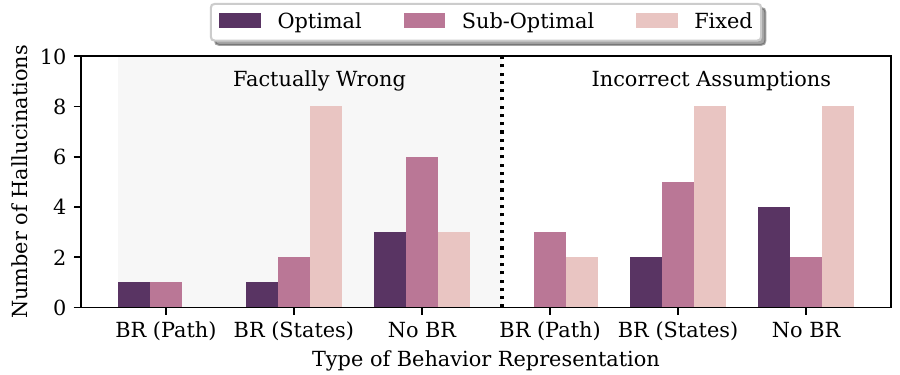}
    \caption{Number of hallucinations that are factually wrong (left) or have incorrect assumptions (right) across our samples.}
    \label{fig:hallucination}
\end{figure}

\subsection{Does Our Approach Decrease Hallucination?}

In order to generate useful explanations, we wish to limit the amount of hallucination exhibited in the LLM's responses.
We quantify the hallucination rate of our proposed approach over 30 generated explanations -- 10 each from an optimal, sub-optimal, and ``fixed'' policy which exhibited state-invariant behavior, i.e. always traveled north.
We compared our explanations with those generated by an LLM when the behavior representation was replaced by a list of state-action pairs randomly sampled from the agent's trajectories $\tau$, and when the behavior representation was removed entirely.
The former represents the case that, rather than producing a compact representation of the agent's behavior, we simply provide direct examples of the behavior and let the LLM reason over them.
While seemingly intuitive, in practice this method does not work well because only a limited number of examples -- 5 in this case -- are able to fit in the LLM's context window.

The results are shown in Fig.~\ref{fig:hallucination} from which we can see that our approach yields significantly fewer numbers of hallucinations than either alternative.
An interesting observation is that the explanations generated without a behavior representation (No BR) actually produces fewer factually wrong hallucinations than a behavior representation consisting of state-action samples (BR (States)) for the ``fixed'' policy.
We find that this is because, in the absence of a behavior representation, the LLM tends to be more conservative in making claims and thus hallucinates less, at the expense of producing less useful explanations.

\section{Conclusion and Future Work}

In this work we propose a model-agnostic framework for producing natural language explanations for an agent's behavior.
Through construction of a \textit{behavior representation}, we are able to prompt an LLM to reason about agent behavior in a way that produces plausible and useful explanations, enables a user to interact and issue follow-up queries, and results in a minimal number of hallucinations, as measured through two participant studies and empirical experiments.
Although we think this is a promising direction, we emphasize that these results are also exploratory in nature and represent first steps. We intend to follow up with more comprehensive studies and analysis in future work.

\newpage

\section*{Acknowledgements}
This work has been funded in part by DARPA under grant HR001120C0036, the Air Force Office of Scientific Research (AFOSR) under grants FA9550-18-1-0251, and the Army Research Laboratory (ARL) under grant W911NF-19-2-0146.

\printbibliography

\end{document}